\documentclass[11pt]{article}
\usepackage{coling2020}
\usepackage{times}
\usepackage{url}
\usepackage{latexsym}
\usepackage{caption}
\usepackage{subcaption}
\usepackage{color,soul}
\usepackage{amsmath}
\usepackage{algorithm}
\usepackage{graphicx}
\usepackage[noend]{algpseudocode}
\usepackage{comment}
\usepackage{url}
\usepackage{multirow}
\colingfinalcopy % Uncomment this line for the final submission
\title{Training Multilingual Machine Translation by \\Alternately Freezing Language-Specific Encoders-Decoders}

\author{Carlos Escolano, Marta R. Costa-juss\`a, Jos\'e A. R. Fonollosa, Mikel Artetxe$\dag$\\
  TALP Research Center, Universitat Polit\`ecnica de Catalunya, Barcelona \\
  $\dag$ IXA NLP Group, University of the Basque Country (UPV/EHU)\\  
  {\tt \{carlos.escolano,marta.ruiz,jose.fonollosa\}@upc.edu}\\{\tt mikel.artetxe@ehu.eus} \\}

\date{}

\begin{document}
\maketitle
\begin{abstract}

 We propose a modular architecture of language-specific encoder-decoders that constitutes a multilingual machine translation system that can be incrementally extended to new languages without the need for retraining the existing system when adding new languages. Differently from previous works, we simultaneously train $N$ languages in all translation directions by alternately freezing encoder or decoder modules, which indirectly forces the system to train in  a common intermediate representation for all languages.  

Experimental results from multilingual machine translation show that we can successfully train this modular architecture improving on the initial languages, while falling slightly behind when adding new languages or doing zero-shot translation. Additional comparison of the quality of sentence representation in the task of natural language inference shows that the alternately freezing training is also beneficial in this direction. %getting average accuracy improvements of 0.5\% . %Overall, we conclude that we are closing the gap between shared and language-specific encoder-decoders. 

% %In this paper we present a modular NMT architecture that allows multilingual translation and incrementally scaling to new languages without retraining any of its components. This achieved by training a multilingual architecture that enforces compatible modules and zero-shot translation. Experimental results show that this can be achieved and retain more than the 92\% of the performance of the bilingual Transformer baselines.
\end{abstract}

\section{Introduction} \label{sec:intro}
Multilingual machine translation generates translations automatically across a number of languages. In the last years, the neural encoder-decoder machine translation architecture \cite{bahdanau2014neural} has allowed for radical improvements in this area. %Thanks to the encoder-decoder architecture, there are viable alternatives to expensive pairwise translation based on classic paradigms\footnote{http://www.euromatrixplus.net}. %\todo{mikel: no entiendo esta footnote, convendria un link, referencia o alguna explicacion}

%The main proposal in this direction is the universal encoder-decoder \cite{johnson2017google} with massive multilingual enhancements \cite{DBLP:journals/corr/abs-1907-05019}. While this approach enables zero-shot translation and is beneficial for low-resource languages, it has multiple drawbacks: (i) the entire system has to be retrained when adding new languages or data; (ii) the quality of translation drops when adding too many languages or for those with the most resources \cite{DBLP:journals/corr/abs-1907-05019}; and (iii) the shared vocabulary grows dramatically when adding a large number of languages (especially when they do not share alphabets). %, and that the entire system has to be retrained when adding new languages or data. 
%Other limitations include the incompatibility of adding multiple modalities such as image or speech. 

Multilingual neural machine translation can refer to translating from one-to-many languages \cite{dong-etal-2015-multi}, from many-to-one \cite{zoph-knight-2016-multi} and many-to-many \cite{johnson2017google}. Within the many-to-many paradigm, existing approaches can be further divided into shared or language-specific encoder-decoders. The latter approaches vary from sharing parameters \cite{firat2016zero,firat2016multi,lu2018neural} to no sharing at all \cite{escolano-etal-2019-bilingual,escolano-etal-2019-multilingual,escolano:2020}. 

%In this paper we are extending research in the line of Escolano et al. 2020 \cite{escolano:2020}, which propose to simultaneously train the $N$ initial languages. Differently, we propose to alternatively train language modules.  

%On the other hand, Escolano et al. (2019) \nocite{escolano-etal-2019-bilingual} propose a new training schedule that allows the system to scale to more languages without modification of the previous components based on joint training and language-independent encoder/decoder modules. However, the architecture requires a multi-way parallel corpus to be trained and it becomes computationally expensive when initially training a large number for language pairs, since all language-independent encoders and decoders have to be trained simultaneously. 

%Beyond multilingual machine translation, there is the necessity to provide automatic systems that are capable of adapting to new topics as interests of the population move on. This is what is known as lifelong learning. While lifelong learning in machine translation has been investigated earlier, the work is basically limited to integrating human post-editions either for bilingual systems \cite{} or low multilingual environments \cite{}. Lifelong learning is also closely related to domain adaptation when new data becomes available and much more in line with previous research in machine translation \cite{} as well as in speech recognition \cite{}, among others. 

Previous research on language-specific encoders-decoders \cite{escolano-etal-2019-bilingual,escolano-etal-2019-multilingual} without sharing any parameters shows that the proposed multilingual system performs quite poorly when new languages are added in the system compared to when languages were jointly trained in the system. One of the main reasons is that the new language module is trained with a frozen module, whereas the initial languages are not trained in such conditions. With respect to the initial architecture, we propose a variation that is specifically designed to add new languages into the system. 

In particular, we propose a new framework that extends previous approach that does not share parameters at all \cite{escolano:2020}. As opposed to any proposal that shares parameters, new languages are naturally added to the system by training a new module coupled with any of the existing ones, while new data can be easily added by retraining only the module for the corresponding language.
% and that can be incrementally extended to new languages (\S\ref{sec:proposed}). 
Our proposal (\S\ref{sec:proposed}) follows \cite{escolano:2020} and it is based on language-specific encoders and decoders that rely on a common intermediate representation space. For that purpose,
 we also simultaneously train the initial \textit{N} languages in all translation directions (\S\ref{sec:background}). However, and differently from \cite{escolano:2020}, we propose to alternately freezing the corresponding encoders and decoders. 
This condition goes beyond teaching each encoder and decoder module to be compatible with the other modules \cite{escolano:2020}. In our new alternately freezing training framework, we are accounting by design that a single model can be improved and extended to new languages as new data become available. %because we are training under the freezing condition. 

We evaluate our proposal on three experimental configurations: translation for the initial languages,  translation when adding a new language, and zero-shot translation (\S\ref{sec:mt}). Our results show that the proposed method is competitive in the first configuration, while still under-performing in the other two conditions. So as to better understand the nature of the learned representations, we run additional experiments on natural language inference, where the alternately freezing outperform in terms of accuracy (\S\ref{sec:xnli}). 

%For both tasks (machine translation and natural language inference), our proposed model shows that it can improve the performance of language-specific architectures without parameter sharing, closing the existent gap with shared encoder-decoders ones. % \todo{mikel: alguna conclusion general para cerrar la intro?}

%\section{Related Work} \label{sec:related}
%\input{sections/related-work.tex}

%\section{Definitions}
%\input{sections/definitions.tex}

%\section{Multilingual training}
%\hl{TO DO}

\section{Background} 
%Multilingual Language-specific Encoders and Decoders} 
\label{sec:background}
In this section, we present the baseline system \cite{escolano:2020} from which our proposed approach (\S\ref{sec:proposed}) follows the idea of training a separate encoder and decoder for each of the $N$ languages available.%, our baseline system is based on previous research \cite{escolano:2020} that is presented in this section. 

%We explore two strategies towards this end: the basic (\S\ref{subsec:basic}) and the frozen (\S\ref{subsec:frozen}) procedures. We do not share any parameter across these modules, which allows to add new languages incrementally without retraining the entire system.

\subsection{Definitions}

We next define the notation that we will be using throughout the paper.
% Before defining the models implemented in this work, we may establish some terminology to aid understanding of the model.
We denote the encoder and the decoder for the $i$th language in the system as $e_i$ and $d_i$, respectively.
% For all experiments, models are referred to as encoder and decoder $e_i$, $d_i$ respectively, for the $i$th language in the system.
For language-specific scenarios, both the encoder and decoder are considered independent modules that can be freely interchanged to work in all translation directions. To refer to the freezing schedules employed in the language-specific models,
each source-target language pair will be described as $f-n$, $n-f$, or $n-n$, where $f$ denotes a frozen language and $n$ a normally trained one.

% each language pair will be defined as a pair $f-n$, $n-f$, or $n-n$, where the leftmost element denotes the source language and where the rightmost one to the target language; each pair it can either have $f$ to indicate a frozen language or $n$ for a normally trained one.

\subsection{Basic Procedure} \label{subsec:basic}

In what follows, we describe the basic procedure presented in \cite{escolano:2020} in two steps: joint training and adding new languages.

\paragraph{Joint Training} The straightforward approach is to train independent encoders and decoders for each language. The main difference from standard pairwise training is that, in this case, there is only one encoder and one decoder for each language, which will be used for all translation directions involving that language. %The training algorithm for this basic procedure is described in Algorithm \ref{alg:train}. 

\begin{algorithm}
\small
\caption{Multilingual training step}\label{alg:train}
\begin{algorithmic}[1]
\Procedure{MultilingualTrainingStep}{}
\State $N \gets \text{Number of languages in the system}$
%\State $S = \{s_{0,0},...,s_{N,N}\} \gets \text{}  \{(e_i,f(d_j)),(f(e_i),d_j),(e_i,d_j)\}$
\State $S = \{s_{0,0},...,s_{N,N}\} \gets \{(e_i,d_j)$ \text{if $i \neq j\}$}
\State $E = \{e_{0},...,e_{N}\} \gets \text{Language-specific encs.}$
\State $D = \{d_{0},...,d_{N}\} \gets \text{Language-specific decs.}$
\For{$i \gets 0$ to $N$}                    
    \For{$j \gets 0$ to $N$}                    
        \If {$s_{i,j} \in S$}
            \State $l_i, l_j = get\_parallel\_batch(i,j)$
            \State $train(s_{i,j}(e_i,d_j),l_i,l_j)$
        \EndIf    
    \EndFor
\EndFor

\EndProcedure

\medskip
\end{algorithmic}
\end{algorithm}

For each translation direction $s_{i,j}$ in the training schedule $S$ with language $i$ as source and language $j$ as target, the system is trained using the language-specific encoder $e_i$ and decoder $d_j$.  
 
\paragraph{Adding New Languages} Once we have our jointly trained model for $N$ languages, the next step is to add new languages. Since parameters are not shared between the independent encoders and decoders, the basic joint training enables the addition of new languages without the need to retrain the existing modules.
%As stated previously no parameter is shared between the different modules of our system and therefore each of them can be employed as a unit without affecting any of the other components.
Let us say we want to add language $N+1$. To do so, we must have parallel data between $N+1$ and any language in the system. For illustration, let us assume that we have $L_{N+1}-L_i$ parallel data. %\todo{mikel: tambien se podria usar $L_i$ de manera mas generica (carlos:hecho)} 
Then, we can set up a new bilingual system with language $L_{N+1}$ as source and language $L_i$ as target. To ensure that the representation produced by this new pair is compatible with the previously jointly trained system, we use the previous $L_i$ decoder ($d_{li}$) as the decoder of the new $L_{N+1}$-$L_{i}$ system and we freeze it. During training, we optimize the cross-entropy between the generated tokens and  $L_i$ reference data but update only the parameters of to the  $L_{N+1}$ encoder ($e_{l_{N+1}}$). %By doing so, we train $e_{l_{N+1}}$ not only to produce good quality translations but also to produce similar representations to the already trained languages. 
Following the same principles, the $L_{N+1}$ decoder can also be trained.
%as a bilingual system by freezing the $L_i$ encoder and training the decoder of the  $L_i-L_{N+1}$ system by optimizing the cross-entropy with the $L_{N+1}$ reference data.

As argued in \cite{escolano:2020}, this approach, even if not sharing parameters at all, does not suffer from the attention mismatch problem \cite{firat2016zero,lu2018neural} because, within an initial set of \textit{N} languages, the system is trained using pair-wise corpus in all translation directions (without requiring multi-parallel corpus as previous works \cite{escolano-etal-2019-bilingual}). Due to the joint training, once the initial system is trained, only $N$ encoders and $N$ decoders ($2*N$) are required. 

\section{Proposed method: Frozen Procedure} \label{sec:proposed}

As discussed in the previous section, new languages are added into the system by learning a new encoder $e$ (or decoder $d$) with a frozen decoder $f(d)$ (or frozen encoder $f(e)$) already in the system. To simulate this condition from the very beginning, we propose modifying the joint training %using what we define as frozen training. %This frozen approach simulates this condition from the very beginning. 
% We propose
by alternately training encoders and decoders while systematically freezing modules. Thanks to this, we are enabling the independent modules to encode and decode using a common representation. For that purpose, we modify Algorithm \ref{alg:train} by adding new training schedules, so line 3 becomes the following:

%\vspace{2mm}

\begin{center}
\small
\hspace{-2mm}3: { 
$S = \{s_{0,0},...,s_{N,N}\} \gets$ $ \text{}  \{(e_i,f(d_j)),(f(e_i),d_j),(e_i,d_j)\}$}
\end{center}
%\vspace{2mm}

Note that we are still training all possible translation combinations among the $N$ languages (avoiding autoencoding and alternating batches in each direction). 

\begin{figure*}[h]
    \centering
    \includegraphics[width=1.01\textwidth]{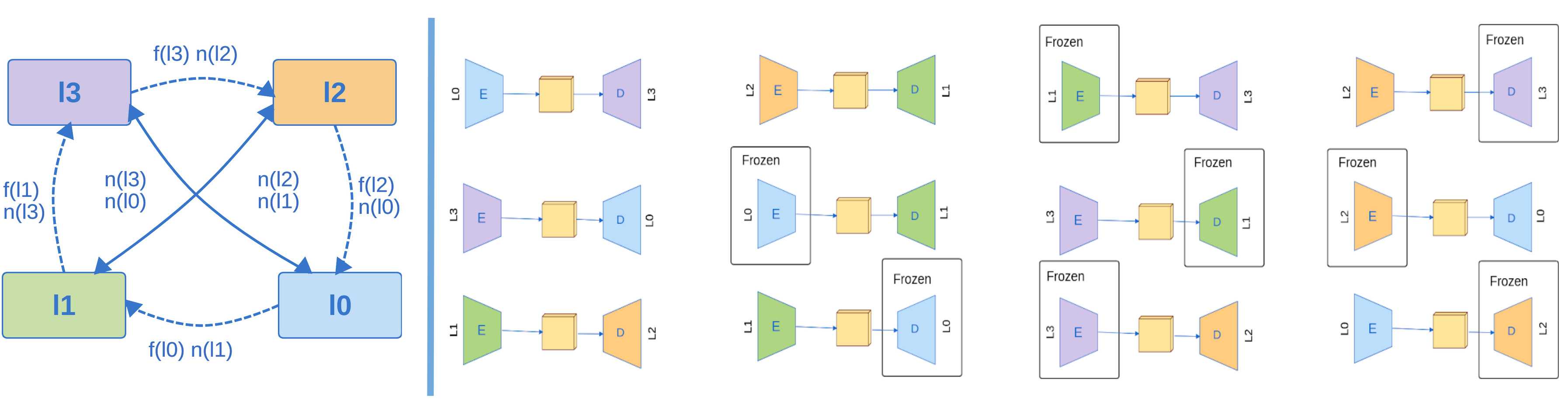}
    \caption{Alternate training of frozen encoders and decoders for 4 languages for particular training schedules: (left) gradient flow, (right) specific freezing modules.}
    \label{fig:trainsch}
\end{figure*}

We freeze the encoder or decoder for a subset of the combinations. When freezing $f(d_j)$, we effectively force the representation of $e_i$ to be as compatible as possible with the representations of the rest of the encoders. This holds true because, if $e_i$ generated an incompatible representation, $f(d_j)$ would be unable to adapt to it given that it is frozen, which would increase their corresponding loss. Similarly, freezing $f(e_i)$ allows $d_j$ to be more robust to a representation that has not been explicitly adapted to it. See 
Figure \ref{fig:trainsch} for an illustration of this with 4 languages for one training schedule.

\begin{comment}

\begin{figure}[h]
    \centering
    \includegraphics[width=0.37\textwidth]{images/trainingschedule.png}
    \caption{Alternate training of frozen encoders and decoders for 4 languages for particular training schedules}
    \label{fig:trainsch}
\end{figure}

\end{comment}

Given a set of N languages, we can define a directed graph where each language is a node and each edge is a training translation direction.  Figure \ref{fig:trainsch} (left) shows the gradient flow between the different languages: $l$ means language; $0...3$ are four different languages in the system; $f()$ means frozen and $n()$ non-frozen, dotted lines means frozen language pairs, continuous lines means non-frozen language pairs. 
Let’s interpret the dotted arrow from box $l_0$ to $l_1$ with the scheme $f(l_0),n(l_1)$. When training the translation direction from language $0$ ($l_0$) to language $1$ ($l_1$), we will freeze the $l_0$ encoder and only update the parameters of the $l_1$ decoder. For the opposite translation direction, we will freeze the $l_0$ decoder and only update the parameters of the $l_1$ encoder. This can be extended to translation pairs: ($l_1$,$l_3$); ($l_2$,$l_3$) and ($l_0$,$l_2$).
For pairs ($l_1$,$l_2$) and ($l_0$,$l_3$) no language is frozen and therefore, there are continuous arrows in two directions because the gradient flows both ways. 

Note that the proposed schedule ensures that for any pair of languages there are two different paths from which information can flow during the training process. This training procedure enables different freezing schedules, which we explore in section \ref{fig:trainsch}. Figure \ref{fig:trainsch} (right) explicitly shows the encoders and decoders that are frozen and non-frozen for 4 languages and the gradient flow represented in the Figure \ref{fig:trainsch} (left).

For N languages, our proposed freezing schedule allows N/2 pairs to be fully trained, while for the rest of the pairs one direction is frozen. The frozen directions (discontinuous lines in Figure \ref{fig:trainsch} (left)) forms an Eulerian cycle between all these languages where one learns from another language and lets another one learning from it. This is a trade-off between freezing and non-freezing that induces the system to jointly learn an intermediate representation.

%The combination of both criteria also creates for each group of 3 languages, a cycle in which two of them are never frozen and the third one leverages their representations, forcing them to the shared representation.

%One of the advantages of this \textit{frozen} training is that it offers a framework for learning at two-levels. The bottom level is alternately training encoders and decoders either frozen or not, and the upper-level controls which pairs have to be frozen. In the section on the experiments we explore prefixed criteria for this upper-level which freezes languages based on language distance \cite{gamallo2017language}. \todo{mikel: este parrafo me suena muy raro. yo lo cambiaria por: ``

%This training procedure enables different freezing schedules, which we explore in sec XX'' %o algo asi} %Closer languages are trained by alternately freezing encoder/decoder while more distant ones are trained without freezing any module.  

This approach is directly extensible to new languages as explained in the previous section for the basic procedure. We can add a new language $N+1$ by training $e_{(N+1)}$ with $f (d_j)$. Once it is trained, this new $e_{(N+1)}$ should be compatible with any $d_j$.

%\section{Data and Implementation}
%\hl{TO DO}
%\input{sections/implementation.tex}

\section{Experiments in Multilingual Machine Translation} \label{sec:mt}

In
%\todo{mikel: yo incluiria ``Experiments in'' o algo asi en el nombre de la seccion}
this section we report machine translation experiments in different settings. 
%Since the main difference between the shared and the language-specific encoders-decoders lies in whether they retrain the entire system when adding new languages, we accordingly design our experiments to compare the systems under this condition. %Additionally, we compare the approaches under initial training (without adding new languages) and in zero-shot translation (when adding new languages and translating between pairs that have not been trained). 

%{\color{red}In addition to the translation quality we report the computational cost of adding a new language in both systems.}
%{\color{red} highlight advantages of frozen}

\subsection{Data and Implementation}
We used 2 million sentences from the \textit{EuroParl} corpus \cite{koehn2005europarl} in German, French, Spanish and English as training data, with parallel sentences among all combinations of these four languages. For Russian-English, we used 1 million training sentences from the \textit{Yandex} corpus\footnote{\url{https://translate.yandex.ru/corpus?lang=en}}. As validation and test set, we used \textit{newstest2012} and \textit{newstest2013} from WMT\footnote{\url{http://www.statmt.org}}, which is multi-parallel across all the above languages. Note that we require multi-parallel corpus in inference (not in training), but only for the specific purpose of evaluating under the condition of zero-shot translation.
%corpus for all languages at hand (German-French-Spanish-English-Russian),  required to test the zero-shot case.
All data were preprocessed using standard Moses scripts \cite{koehn2007moses}

We evaluate our approach in 3 different settings: (i) the \textit{initial} training, covering all combinations of German, French, Spanish and English; (ii) \textit{adding} new languages, tested with Russian-English in both directions; and (iii) \textit{zero-shot} translation, covering all combinations between Russian and the rest of the languages.
% With these data, we proposed three experimental configurations. One is what we call \textit{initial} training, which includes training German, French, Spanish and English and all possible language pairs, since we had parallel data for all of them. The configuration was \textit{adding} new languages, which consisted of adding Russian-English in both directions. Finally, the \textit{zero-shot} option, which consists in translating from Russian to German,Spanish,English and in the opposite directions with no training data provided.
\label{sec:mnmt}

Experiments were done using the Transformer implementation provided by Fairseq\footnote{Release v0.6.0 available at \url{https://github.com/pytorch/fairseq}}.
We used 6 layers, each with 8 attention heads,
% We used 6 blocks of multihead attention blocks of 8 heads each, 
% embedding/hidden dimensionality
an embedding size of 512 dimensions,
and a vocabulary size of 32k subword tokens with 
% of 512 and fixed learning rate of 0.001 and vocabulary size of 32 thousand
Byte Pair Encoding \cite{sennrich-etal-2016-neural} (per pair). Dropout was 0.3 and trained with an effective batch size of 32k tokens for approximately 200k updates, using the validation loss for early stopping. In all cases, we used Adam \cite{kingma2014adam} as the optimizer, with learning rate of 0.001 and 4000 warmup steps. %\todo{mikel: igual se puede poner lo del decay entonces} 
All experiments were performed on an NVIDIA Titan X GPU with 12 GB of memory. % The joint training was performed on an NVIDIA Titan X GPU with 12 GB of RAM as was the addition of new languages to the trained system. %Initial training lasted for approximately a week in all configurations while the new languages were added in approximately one day of training.

\subsection{Basic vs Frozen}
\label{sec:expfrozen}

We used multiple configurations for the case of frozen. The main difference in these configurations is which languages and modules are frozen. %Basically, we follow the criteria of language distance proposed by Gamallo et al. \citet{gamallo:2017}. 

Basically, we followed  the criterion of linguistic families. We assumed that languages that belong to the same linguistic family are closer than those that are not. 

Based on this, English/German-French/Spanish are candidates to be farthest language pairs. To decide which of Spanish and French is the farthest from English,  we follow the criterion of language distance proposed by \cite{gamallo2017language}; We used  an  
%Distances between language pairs is 
%reported in table \ref{tab:trainingsch}. 
English-Spanish distance of 18 which is higher than the English-French distance (16). Therefore, we chose as more distant pairs of languages English-Spanish and German-French. The reason to choose German-French is so that the same language will not be frozen in several pairs, otherwise, we risk that this language will not able to train well for all translation directions. English-German and French-Spanish are chosen as  the closest language pairs. 

We could either train without freezing the farthest pairs (\textit{far}) or train without freezing the closest pairs of modules (\textit{close}). The order of the modules to be frozen is given in Table \ref{tab:baselines}. %\todo{mikel: todo este parrafo se me hace un tanto enrevesado. yo definiria que idiomas cubren las dos configuraciones, y el criterio que se ha usado para ello puede ir en una footnote}

We found the option of freezing in all cases to perform poorly in our preliminary experiments. Finally, and inspired by curriculum learning techniques \cite{curriculumlearning}, we can use an adaptive scheme (\textit{adapt}) which basically after each epoch computes a new training schedule based on the average validation loss. Two pairs with higher loss are not frozen, and then, the rest of the languages are frozen composing the Eulerian cicle as explained in section \ref{sec:proposed}. %and leave the exploration of alternative freezing schedules for future work.
% The option of freezing in all cases is not reported because BLEU results were quite low, even if translations were somehow understandable. There are many alternative training schedules that would be worth testing in future work. 

\paragraph{Initial Training} Table \ref{tab:baselines} shows a comparison between training language-specific encoders and decoders  using either the basic or the frozen procedure. 
The frozen procedure clearly outperforms the basic one for all language pairs. When comparing different training schedules for the frozen procedure (\textit{far} vs \textit{close}), 
the results show that the \textit{far} configuration in the frozen procedure is the best one, except in 4 (out of 12) language directions. When comparing \textit{far} with \textit{adapt}, the former is again the best one, except in 4 (out of 12) language directions. Adapting the training schedule based on the loss ends up as a pre-defined one. What we observed in the adaptive schedule is that we start from the far configuration and the system moves to the close configuration in one iteration and does not change anymore. When using the loss as a criteria, we are introducing external factors, so this criteria does not seem flexible enough.
Given these results, the rest of the experiments were performed using the \textit{far} training schedule.

%{\color{red} All 4 of these pairs involve to not frozen pair German-English that is the one that benefits more from the difference in schedule. This may be result of the difference in distance in the not frozen pairs, being English more similar to German than French in the far configuration.}
%\todo{mikel: estaria bien si pudieramos dar una explicacion, aunque sea especulativa} 

%In those cases the shared comparable model performs better than the language independent ones. However for the pairs the frozen model when the furthest languages are never frozen, the model outperforms the shared one for all 4 translation directions.  %Therefore, we test the same version (far) but inverting the module to be freezed (encoder and decoder) and this is the spin version. This option does not improve.  

\begin{table*}[]
\begin{minipage}{.5\textwidth}
\centering
\begin{tabular}{|l|c|cc|cc|c|}
\hline
      & Basic  & \multicolumn{5}{c|}{Frozen} \\ \hline
      &&\multicolumn{2}{c|}{Far} & \multicolumn{2}{c|}{Close} & Adapt \\ \hline
     % &&& BLEU & schedule & BLEU &schedule & \\ \hline
de-en & 22,04 & 23,25 & n-f & 23,47 &n-n & \textbf{24,06}  \\
de-es & 22,38 & \textbf{24,25} & n-f &23,02 &f-n & 23,78\\
de-fr & 22,57 & \textbf{25,08} &n-n & 23,46 & n-f & 24,00 \\
en-de & 19,44 & 19,55 & f-n &  20,88 &n-n & \textbf{21,30}\\
en-es & 26,79 & \textbf{{28,6}} &n-n   & 27,42& n-f & 27,97   \\
en-fr  & 26,94 & 27,72 & f-n&  \bf 28,03 &f-n & 27,81 \\
es-de & 17,7  & 18,21 & f-n&  \textbf{18,44} &n-f & 18,43 \\
es-en & 24,9 & \textbf{{27,06}} &n-n & 25,3 &f-n    & 26,35\\
es-fr & 27,31  & 29,34& f-n & 28,93 &n-n & \bf 29,92\\
fr-de & 16,88 & \textbf{19,22} &n-n & 17,19 &f-n & 18,34 \\
fr-en & 23,5 & \bf 25,11 &n-f  & 24,31 &n-f & 24,91\\
fr-es & 26,78 & \bf 28,14 & f-n & 27,31 &n-n & 28,08\\ \hline
\end{tabular}
\caption{\label{tab:baselines} Initial training. In bold, best global results.}
\end{minipage}
\begin{minipage}{.5\textwidth}
%\begin{table}[]
\centering
\begin{tabular}{|l|cc|}
\hline
      & Basic & Frozen \\ \hline
ru-en & \textbf{{ 25,52}} & 25,08  \\
en-ru & \textbf{{21,44}} & 21,33  \\ \hline
\end{tabular}
\caption{\label{tab:add} Adding a new language translation.}
%\end{table}
%\begin{table}[]
\vspace{10mm}
\centering
\begin{tabular}{|l|cc|}
\hline
   & Basic & Frozen \\ \hline
ru-de & \bf 12,73 & 11,85  \\
ru-es & \bf 18,71 & 15,31  \\
ru-fr  & \bf 18,05 & 17,46  \\
de-ru & 14,39 & \bf 14,99  \\
es-ru & \bf 15,93 & 14,85  \\
fr-ru &  \bf 15,1 & 14,99  \\ \hline
%de-de & auto-enc. & \bf 52,07 &     45,48 &	\it 45,94  \\
%en-en & auto-enc. & \bf 71,01 &	\it    61,92 &	51,95  \\
%es-es & auto-enc, & \bf 64,33 &	    61,02 &	\it 62,28  \\
%fr-fr & auto-enc. & \bf 64,4  &	  \it  59,73 &	58,27  \\
%ru-ru & auto-enc. & \bf 55,13 &	    10,56 &	\it 15,3   \\ \hline
\end{tabular}
\caption{\label{tab:zero}Zero-shot translation.}
%\end{table}
\end{minipage}
\end{table*}

%We take the far configuration from the frozen procedure and we compare it to the basic procedure in two more conditions. 

\paragraph{Adding New Languages} As shown in Table \ref{tab:add}, the basic procedure performs slightly better than the frozen one when adding a new language into the system.% When adding a new language to the system, training it with a module already in the system (ru-en), the basic procedure outperforms the frozen one. See Table \ref{tab:add}. 

\paragraph{Zero-shot} As shown in Table \ref{tab:zero}, the basic procedure also outperforms the frozen procedure in zero-shot translation for 6 (out of 7) translation directions. 

%Second, when doing zero-shot translation. Results are shown in Table \ref{tab:zero}. For these two conditions, the conclusions are reverted and basic procedure outperfoms the frozen procedure in 6 (out of 7) language directions.

%{\color{red} comment on auto-encoder}

\
\subsection{Discussion}

Surprisingly, we are obtaining improvements in the condition that we did not expect (the \textit{initial} one). The very first motivation of the frozen procedure is to simulate the conditions of adding a new language which consists in training with a frozen module. So, we were expecting the frozen procedure to improve over the basic one when \textit{adding} new languages and performing \textit{zero-shot} translation. Under these two conditions, the frozen procedure is slightly worse than the basic procedure. 

However, and proven by the nice improvements reported under the \textit{initial} condition, we believe that the alternatively freezing scheme that we are proposing has a big potential to be exploited in investigating alternative training schedules, e.g. employing more advanced techniques of curriculum learning \cite{curriculumlearning}.  %instead of the adding new languages for which we trained for),

Taking a closer look at the results, we can find different behaviors between languages pairs frozen during training and those that not. Starting by the initial condition, non-frozen pairs (German-French, Spanish-English) are the best performing systems and further outperform the basic procedure by an average of 2,21 BLEU points, while frozen ones (German-English, German-Spanish, French-English) reduce those gains to 1,19 points. Those results show that the additional information available to not frozen language-pairs is beneficial in the initial task. The additional information comes from the fact that both encoder and decoder are updated during training in the case of German-French and Spanish-English.

On the other hand, this behaviour is reversed when looking at zero-shot results. %Excluding the pair that outperforms the baseline system (German-Russian), 
In the case of zero-shot, Spanish that was not frozen to English is the worst performing language in the zero-shot task. Whereas languages (French, German) that were frozen to English are closer to the baseline performance (even better in the German-to-Russian case). %Since Spanish-English is not frozen during training, this language pair is trained with more information than German-English or French-English which are frozen. 
The additional information in training, which was beneficial for the initial task, seems to be detrimental for zero-shot translation. %difference may be caused by the imbalance on the amount of information that this language pair (Spanish-English) has during training compared to the other pairs (German-English, French-English). %In any other aspect, those pairs are trained as the Spanish ones.
Previous works \cite{lu2018neural,gu-etal-2019-improved} reported similar conclusions, showing that language-specific information (or spurious correlation) harms the model's generalization to other languages as the task of zero-shot requires.

To further understand the performance of the frozen procedure, we are exploring its performance in the task of natural language inference in the next section.

%{\color{red} comment in discarded methods: everything frozed}

\section{Experiments in Cross-lingual Natural Language Inference} \label{sec:xnli}

Given
%\todo{mikel: yo incluiria ``Experiments in'' o algo asi en el nombre de la seccion}
two sentences, a reference and a hypothesis,  the natural language inference (NLI) task consists in deciding whether the relationship between them is \textit{entailment}, \textit{contradiction} or \textit{neutral}. This task has been addressed as a classification problem using the relatedness of the representation of sentences. %This task is interesting in our case because we can compare how useful  the encoder representation is in an alternative downstream application. 
Following the procedure of \cite{conneau-etal-2018-xnli} a model is trained to classify a classifier to task. In the original work, the model consisted of a bidirectional recurrent encoder and  as classifier, two fully connected layers with ReLU and Softmax activation respectively. The classifier is fed with the following combination of the encoding of both reference and hypothesis:
\begin{equation}
    \label{eq:encoding}
	h = [u,v,|u-v|,u*v]
\end{equation}
where $u$ is the reference encoding, $v$ is the hypothesis encoding and $*$ is the element multiplication of both vector representations. 
In that work, encoders were trained specifically on the task of natural language inference, independently for each language and  representations were forced to share representation space by means of additional loss terms.
For our task, we want to study the shared space already trained by the different configurations of multilingual machine translation systems from section \ref{sec:mnmt}.  For each of them, a classifier is trained using its English encoder, which is frozen to help the classifier learn from the current shared space. To keep the encoding as described in equation \ref{eq:encoding} while using a Transformer encoder, the contextual embeddings are averaged to create a fixed-sized sentence representation. This approach was previously proposed by \cite{DBLP:journals/corr/abs-1907-05019}, where pooling was employed and allows the representation size to be fixed while not adding extra padding to the data; at the cost of producing an information bottleneck for the classification because all sentence information has to be condensed into a single fixed-size vector, independently of the sentence's length.

%{\color{red}This techniques has been previously employed to analyze the share representation created by  multilingual systems \cite{poliak-etal-2018-collecting-diverse,artetxe2019massively}. In this work we want to compare the representations generated by the proposed systems without any additional adaptation to the NLI task.}

Given that all language pairs in both language-specific architectures were trained to share sentence representations, we can evaluate the classifier's performance compared with that of all the other languages in the multilingual system 
without any extra adaptation.%, as the models were trained jointly for all language pairs to share sentence representations.   
%\todo[inline]{mikel: toda esta intro se me hace excesivamente largo y enrevesado. creo que seria suficiente con decir que xnli es ampliamente usado para evaluar representaciones multilingues, y en nuestro caso la usaremos para entender mejor el comportamiento de los distintos sistemas. todos los detalles del clasificador pueden ir en el 5.1.}

\begin{figure}
\begin{minipage}{.5\textwidth}
    \includegraphics[scale=0.15]{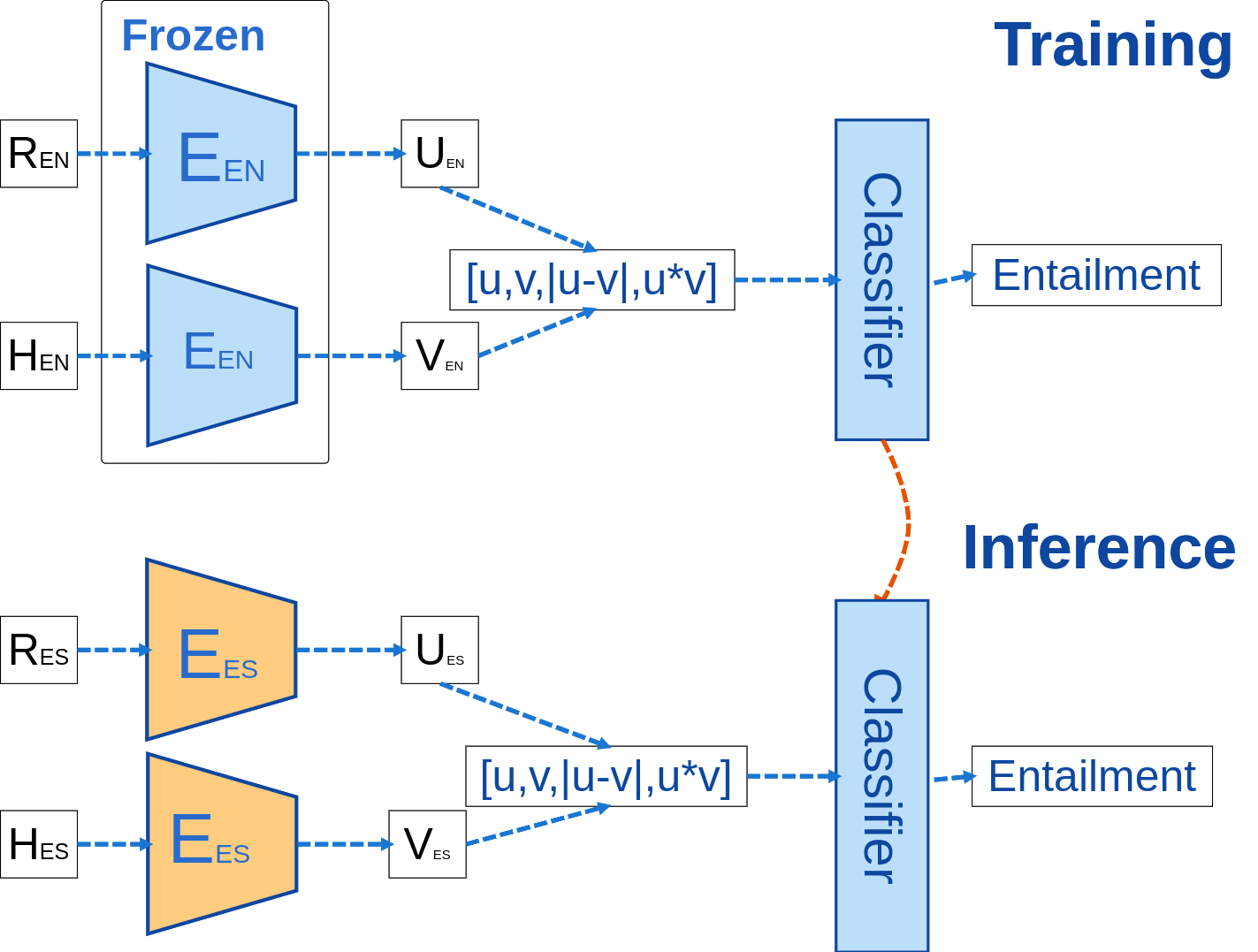}
    {\captionof{figure}[]{Experiment setup for NLI.}}
\end{minipage}
\begin{minipage}{.5\textwidth}
\begin{center}
   \vspace{1.2cm}
   \begin{tabular}{l|cc|}
            \cline{2-3}
                         & Basic & Frozen        \\ \hline
            %\multicolumn{1}{|l|}{en$^s$}                           & 71.6  & \textbf{{73.9}} \\ \hline
            \multicolumn{1}{|l|}{en}                        & 58.1  & \textbf{{60.2}} \\ \hline
            \multicolumn{1}{|l|}{de}                  & 54.7  & \textbf{57.8}          \\
            \multicolumn{1}{|l|}{es}             & 54.6  & \bf 55.6          \\
            \multicolumn{1}{|l|}{fr}                      & 55.0  & \bf 57.9          \\
            \multicolumn{1}{|l|}{ru}            & 37.6  & \textbf{41.4}            \\ \hline
        \end{tabular}
        \vspace{1.8cm}
        {\captionof{table}[]{NLI (en, de, es, fr, ru) results. In bold best results. \label{tab:xnli}}}
\end{center}
\end{minipage}
\end{figure}

\subsection{Data and implementation}

For this task, we use the MultiNLI corpus \footnote{https://cims.nyu.edu/~sbowman/multinli/} for training, which contains approximately 430k entries. %We exclude all segments without golden truth (around 9k and 200, respectively). %\todo{mikel: para la proxima, lo comun es usar XNLI junto a MultiNLI, y no SNLI} 
We use the XNLI validation and test set \cite{conneau-etal-2018-xnli} for cross-lingual results, which contain 2.5k and 5k segments, respectively, for each language. 

We use the exact same encoders trained for the machine translation experiments (\S\ref{sec:mnmt}), which are \textit{not} further retrained or fine-tuned for this task.
% Encoders are the same ones trained in the machine translation experiments from Section \ref{sec:mnmt} without retraining for this task.
%We train a classifier with 128 hidden units on top of the encoder using English labeled data, and evaluate the zero-shot transfer performance in the rest of the languages.
%\todo{mikel: la descripcion mas detallada del clasificador puede ir aqui en lugar de en la intro}
 A classifier with 128 hidden units is  exclusively trained on top of the English encoder, which is the only language for which we have training data. 
Note that %the shared encoder-decoder, shares vocabulary and parameters among languages within the same encoder, which is the same for all languages and is the one that has been used to train the classifier with the English data; in contrast, 
the basic and frozen language-specific encoder-decoders, each language has its own encoder, and both vocabulary and parameters are fully independent. 
%The classification task is exclusively trained on the English encoder, which is the only language for which we have training data. 

%Then, in the case of the language-specific architecture, other languages (German, Spanish, French and Russian) use their corresponding language-specific encoder together with the classifier layers previously trained (only in English). In the case of the shared architecture, the same encoder is used together with the classifier layers.  

\subsection{Results}

Table \ref{tab:xnli} shows the results for the XNLI task using either basic or frozen language-specific encoder-decoders. Note that our goal is not to improve the state-of-the-art in this task, but rather to analyze the nature and quality of the cross-lingual representations arising in the different multilingual architectures. %\todo{mikel: esto ultimo tambien podria ir en la intro}

% Table \ref{tab:xnli} shows the results for both SNLI and XNLI tasks using either shared or language-specific encoder-decoders. We are comparing the multilingual models on the monolingual task of natural language inference (NLI, English) and the task of cross-lingual natural language inference for German, Spanish, French and Russian (XNLI). The results are below \cite{conneau-etal-2018-xnli}, because our focus is on testing the relatedness of the obtained representations, not the design of suited systems for the task, as in the original work were models were trained for the task using cross-lingual embeddings. 

%Neither the shared nor the language-specific architectures have cross-lingual embeddings and none of the encoders are specifically trained for the NLI task. However, the point of these experiments is to compare the multilingual architectures (shared vs language-specific) and their corresponding variations (Shared, Shared$^{RU}$, Basic, Frozen) on their ability to produce useful and generalizable sentence representations.

\paragraph{Basic vs Frozen} The frozen procedure outperforms the basic one for all cases. %For the XNLI task (de, es, fr, ru), the basic and frozen procedures show equivalent accuracies. Basic is better for Spanish and French, while frozen is better for German and Russian. 
This shows that the frozen procedure helps in creating a coherent intermediate representation.

%Note that we are obtaining the worst results in the case of Spanish, which is coherent with the worst results obtained for zero-shot MT. These results enforce the hypothesis that providing additional language-specific information to some pair during training may be harming their generalization to other languages.%, as shown by the Spanish results both here in NLI and, previously, in zero-shot MT. 
% Again, coherently with the zero-shot results, the next worse language is French, %frozen system also underperforms the basic one, 
%which may be explained by the fact that either French is the closest language to Spanish.% and/or the particular frozen training scheme that we are following.

%This may show that even though the basic configuration's better information flow may help on the task of machine translation, freezing the components may produce a better generalization that can be exploited for downstream tasks.

\section{Visualization}

%congelado

%OLD
%70 votos para lograr la presidencia
%se pudo en parte limitar los daños
%cancer of the breast
%it is also a risk factor
%ses restriccions ne son pas

%new
%recomiendo que se haga el test
%parte el sistema americano
%the American democratic system
%they were able to partially limit the damage
%have adopted laws requiring

%google
%prevenir la enfermedad
%take the test or not
%los productos
%it is also a risk factor
%ses restriccions ne son pas 

\begin{figure}[h!]
\begin{minipage}{.5\textwidth}
\centering
    \centering
    \includegraphics[width=1\textwidth]{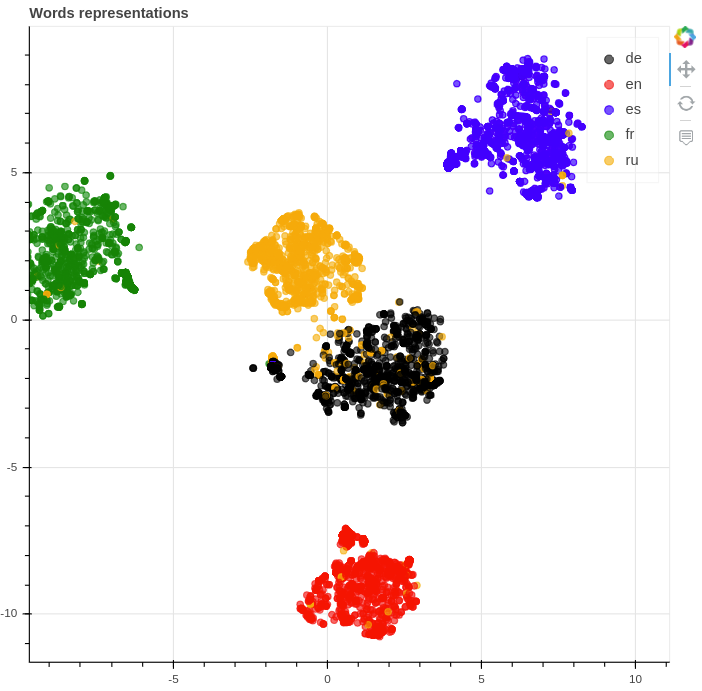}
\end{minipage}
\begin{minipage}{.5\textwidth}
%\begin{figure}[h!]
    \centering
    \includegraphics[width=1\textwidth]{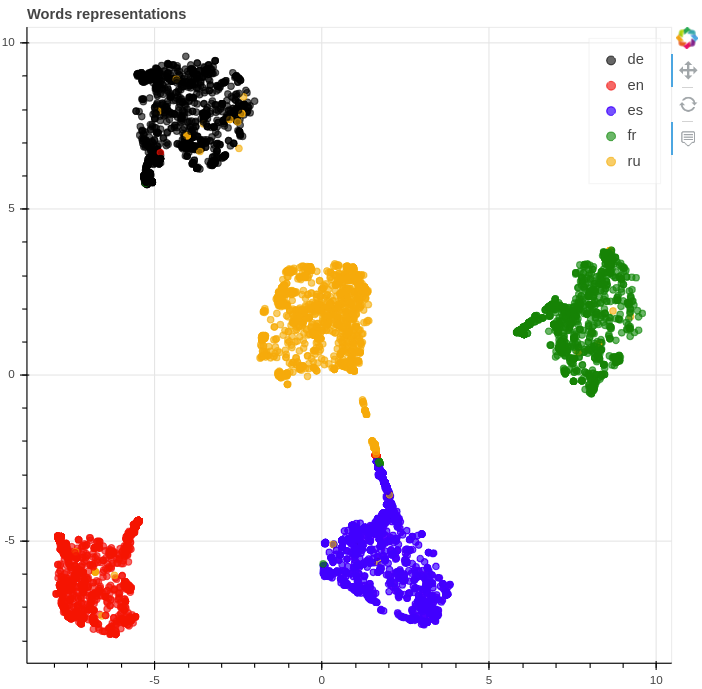}
%\end{figure}
\end{minipage}
\caption{Visualization of words embeddings for the basic (left) and frozen (right) approaches.}
    \label{fig:vis1}
\end{figure}

\begin{figure}[h!]
\begin{minipage}{.5\textwidth}
\centering
    \centering
    \includegraphics[width=1\textwidth]{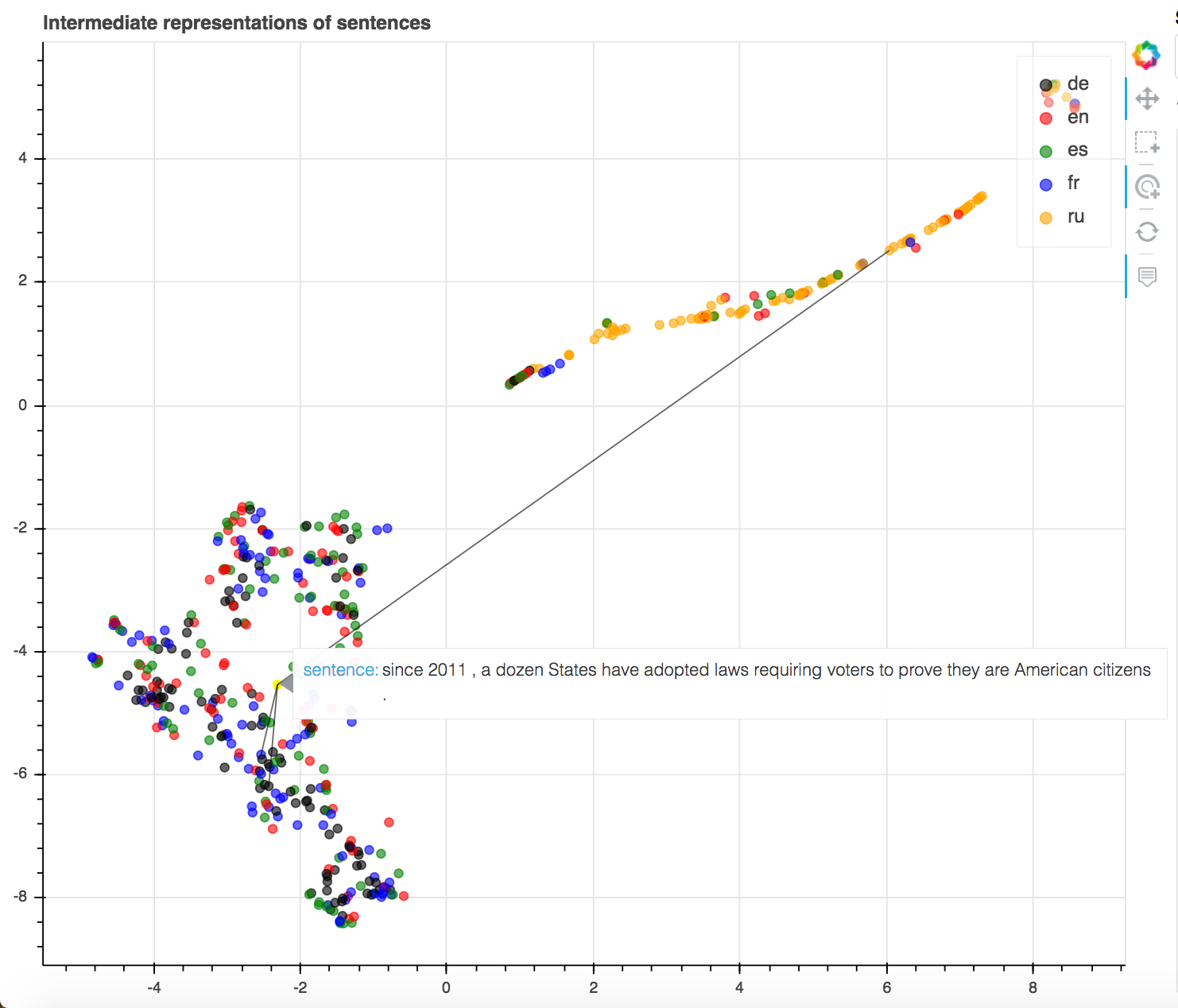}
\end{minipage}
\begin{minipage}{.5\textwidth}
%\begin{figure}[h!]
    \centering
    \includegraphics[width=1\textwidth]{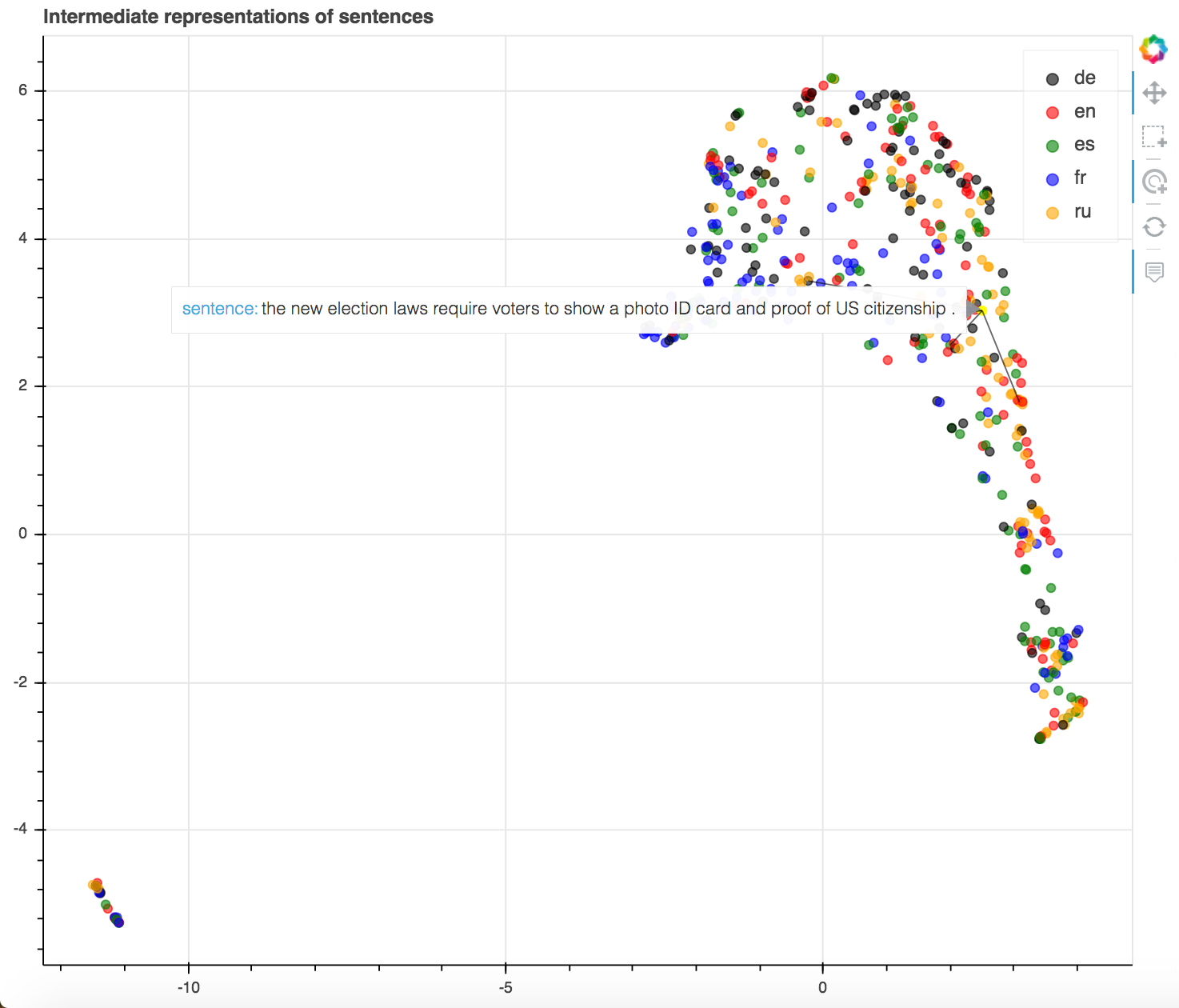}
%\end{figure}
\end{minipage}
    \caption{\label{fig:vis2}Visualization of sentence representations for the basic (left) and frozen (right) approaches.}
\end{figure}

%\section{Sentence Embedding Visualization}

In what follows, we use a tool \cite{escolano-etal-2019-multilingual} freely available\footnote{https://github.com/elorala/interlingua-visualization}, that allows us to visualize intermediate sentence representations. The tool uses the encoder's output fixed-representations as input data and performs a dimensionality reduction in these data using UMAP \cite{mcinnes2018umap-software}. Additionally, we are also adding a comparison of words embeddings. In both cases, we make the comparison for the two architectures (basic and frozen) from the paper. 

Figure \ref{fig:vis2} shows the intermediate representation of 100 sentences in each languages (German, English, Spanish, French and Russian) %for the 3 configurations under study (shared, basic and frozen). 
 of the sentence \textit{since 2011 a dozen States have adopted laws requiring voters to prove they are American citizens} and the corresponding translations. %Figure \ref{fig:vis2} shows the representation for sentence \textit{en ese sentido, estas medidas minar\'an en parte el sistema democrático americano}. 

Note that, although all the points seems to be mixed together, the representation of the same sentence in different languages is not placed exactly in the same point in the space. For the example, representation seems closer between sentences for the frozen architecture than in the basic case. In the case of the basic architecture, the sentence representations of  Russian , which is the language that is added later, seem to be quite far from those of the others languages. The frozen training schedule, while forcing the language-specific modules to adapt to the representations already learned from other languages may help producing a more general language representation, as already shown in the downstream task experiments. 

While sentence representation are not fully shared between languages,  our next question is if that difference in the representation is also present in the multilingual word representations. Figure \ref{fig:vis1} shows a two-dimensional  representation of the words embeddings of each architecture; obtained using UMAP dimensionality reduction, as done for sentences in Figure \ref{fig:vis2}. 

Visualizations shows similar representations for word embeddings, in the sense that each language has its own cluster, which indicates that each of the modules has learnt its individual word representation. This finding shows that while a shared word embedding space might not be mandatory for the task it would be interesting to explore its impact in future work. Table \ref{tab:example} shows how for the sentence \textit{the new election laws require voters to show a photo ID card and proof of US citizenship}, the basic architectures wrongly translate \textit{photo ID} while the frozen architecture is able to keep the original term.

% Please add the following required packages to your document preamble:
% \usepackage{multirow}
\begin{table*}[h!]
\small
\begin{tabular}{|p{1.5cm}|p{1.5cm}|p{12cm}|}
\hline
\textbf{System}         & \textbf{Languages} & \textbf{Sentence}                                                                                                                             \\ \hline
Reference               &                    & the new election laws require voters to show a photo ID card and proof of US citizenship                                                     \\ \hline
\multirow{4}{*}{Basic}  & DE                 & the new electoral laws require the voters to present a ray of light and evidence of US citizenship                                          \\
                        & ES                 & the new electoral laws require voters to present an identity document with a photograph , as well as a test of American citizenship          \\
                        & FR                 & the new electoral laws require voters to present an identity card with images and evidence of American citizenship                           \\
                        & RU                 & the new electoral laws require the voter to have a photograph card and proof of American citizenship                                         \\ \hline
\multirow{4}{*}{Frozen} & DE                 & the new electoral laws require the voters to produce a beacon and proof of US citizenship                                                    \\
                        & ES                 & the new electoral laws require the voters to submit an identity card with a photograph as well as a test of American citizenship             \\
                        & FR                 & the new electoral laws require the voters to present an identity card with a photograph and proof of American citizenship                    \\
                        & RU                 & the new electoral laws require the voters to have a photo ID and proof of the existence of American citizenship                              \\ \hline
\end{tabular}
\caption{\label{tab:example}Translation examples for each of the tested architectures: basic and frozen.}
\end{table*}

%\begin{figure}
%    \centering
%    \includegraphics[width=0.5\textwidth]{images/google2.png}
%     \includegraphics[width=0.5\textwidth]{images/singcongelar2.png}
%    \includegraphics[width=0.5\textwidth]{images/congelado2.png}
%    \caption{Visualization 2 for the shared (up), basic (middle) and frozen (bottom) approaches.}
%    \label{fig:vis2}
%\end{figure}

Note that the visualization and the example shown in here only pretends to provide a small qualitative and interpretable analysis of the proposed models.

%\section{Discussion} \label{sec:discussion}

%\input{sections/discussion.tex}

\section{Conclusions} \label{sec:conclusions}

%Por un lado, habria que justificar que se han hecho los 
%experimentos con un numero reducido de idiomas por el coste 
%computacional, que seria interesante comprobar como se comportan los 
%distintos sistemas con un alto numero de idiomas como future work o 
%algo asi. 

%combinar ambas arquitecturas?

In this paper, we present a novel training methodology language-specific encoders-decoders specifically designed to perform well when incrementally adding new languages into the system without having to retrain it. We believe that this approach can be particularly useful for situations in which a rapid extension of an existing machine translation system is critical (e.g. aiding in a disaster crisis where international help is required in small region, or developing a translation system for a client). 
%When adding a new language, the language-specific encoder-decoders outperform the shared ones by 0.9 BLEU on average and, most importantly, the training of this new language was done in only one day, as opposed to the week taken by the shared system. 

We provide a comparison of the language-specific procedure for the previously proposed basic procedure \cite{escolano:2020} and our current proposal of alternatively freezing encoders-decoders during training. Comparison is done for the task of machine translation in the situation of \textit{initial} language training, \textit{adding} new languages and \textit{zero-shot} translation. Performance of the frozen procedure is better in the first situation and slightly worse in the other two. Additional experiments in natural language inference show that the accuracy of classification when using the sentence representation of the language-specific encoders is higher in the current proposed frozen procedure. 

While we obtained improvement in the machine translation condition that we did not expect (the \textit{initial} one instead of the \textit{adding} new languages for which we trained for), we believe that the alternatively freezing scheme has further potential to be exploited in investigating alternative training schedules. %For this, we will consider an upper-level of learning \cite{graves:2017} and tentatively starting with a pre-fixed criterion such as the one we have implemented of language distances \cite{gamallo2017language} and learning which modules to freeze based on this. 

% and lower for the cross-lingual case (German, Spanish, French and Russian).% but it slightly improves in English (by 0.2). 
%Also, we observe that the difference in accuracy for those worse is only 3 points, in average, for the initial languages (German, Spanish and French), 10.8 points for the added language (Russian).

%Similar to \citet{DBLP:journals/corr/abs-1907-05019}, our results suggest that the shared architecture is beneficial for languages share the same script thanks to the joint vocabulary. In the future, we would like to further compare the shared and language-specific encoders-decoders in cases where the languages do not share scripts (e.g. Chinese, Arabic, Russian and Greek). If this hypothesis is true, we could further improve the language-specific encoder-decoders by using strategies that mitigate the disadvantage of disjoint vocabulary. A visual comparison of the multilingual word and sentence embeddings provided by each of the presented approaches is shown in the appendix.

\section*{Acknowledgments}

%This work is supported in part by the bla bla bla

Google Faculty Research Award 2018 and 2019. Authors want to specially thank Oriol Vinyals for helpful and inspiring discussions. This work is also partially funded by the Spanish Ministerio de Econom\'ia y Competitividad, the European Regional Development Fund through the  postdoctoral senior grant Ram\'on y Cajal and by the Agencia  Estatal  de  Investigaci\'on through the project EUR2019-103819.

\bibliography{emnlp-ijcnlp-2019}

\begin{thebibliography}{}

\bibitem[\protect\citename{Arivazhagan \bgroup et al.\egroup
  }2019]{DBLP:journals/corr/abs-1907-05019}
Naveen Arivazhagan, Ankur Bapna, Orhan Firat, Dmitry Lepikhin, Melvin Johnson,
  Maxim Krikun, Mia~Xu Chen, Yuan Cao, George Foster, Colin Cherry, Wolfgang
  Macherey, Zhifeng Chen, and Yonghui Wu.
\newblock 2019.
\newblock Massively multilingual neural machine translation in the wild:
  Findings and challenges.
\newblock {\em CoRR}, abs/1907.05019.

\bibitem[\protect\citename{Bahdanau \bgroup et al.\egroup
  }2014]{bahdanau2014neural}
Dzmitry Bahdanau, Kyunghyun Cho, and Yoshua Bengio.
\newblock 2014.
\newblock Neural machine translation by jointly learning to align and
  translate.
\newblock {\em arXiv preprint arXiv:1409.0473}.

\bibitem[\protect\citename{Bengio \bgroup et al.\egroup
  }2009]{curriculumlearning}
Yoshua Bengio, J\'{e}r\^{o}me Louradour, Ronan Collobert, and Jason Weston.
\newblock 2009.
\newblock Curriculum learning.
\newblock In {\em Proceedings of the 26th Annual International Conference on
  Machine Learning}, ICML ’09, page 41–48, New York, NY, USA. Association
  for Computing Machinery.

\bibitem[\protect\citename{Conneau \bgroup et al.\egroup
  }2018]{conneau-etal-2018-xnli}
Alexis Conneau, Ruty Rinott, Guillaume Lample, Adina Williams, Samuel Bowman,
  Holger Schwenk, and Veselin Stoyanov.
\newblock 2018.
\newblock {XNLI}: Evaluating cross-lingual sentence representations.
\newblock In {\em Proceedings of the 2018 Conference on Empirical Methods in
  Natural Language Processing}, pages 2475--2485, Brussels, Belgium,
  October-November. Association for Computational Linguistics.

\bibitem[\protect\citename{Dong \bgroup et al.\egroup
  }2015]{dong-etal-2015-multi}
Daxiang Dong, Hua Wu, Wei He, Dianhai Yu, and Haifeng Wang.
\newblock 2015.
\newblock Multi-task learning for multiple language translation.
\newblock In {\em Proceedings of the 53rd Annual Meeting of the Association for
  Computational Linguistics and the 7th International Joint Conference on
  Natural Language Processing (Volume 1: Long Papers)}, pages 1723--1732,
  Beijing, China, July. Association for Computational Linguistics.

\bibitem[\protect\citename{Escolano \bgroup et al.\egroup
  }2019a]{escolano-etal-2019-bilingual}
Carlos Escolano, Marta~R. Costa-juss{\`a}, and Jos{\'e} A.~R. Fonollosa.
\newblock 2019a.
\newblock From bilingual to multilingual neural machine translation by
  incremental training.
\newblock In {\em Proceedings of the 57th Annual Meeting of the Association for
  Computational Linguistics: Student Research Workshop}, pages 236--242,
  Florence, Italy, July. Association for Computational Linguistics.

\bibitem[\protect\citename{Escolano \bgroup et al.\egroup
  }2019b]{escolano-etal-2019-multilingual}
Carlos Escolano, Marta~R. Costa-juss{\`a}, Elora Lacroux, and Pere-Pau
  V{\'a}zquez.
\newblock 2019b.
\newblock Multilingual, multi-scale and multi-layer visualization of
  intermediate representations.
\newblock In {\em Proceedings of the 2019 Conference on Empirical Methods in
  Natural Language Processing and the 9th International Joint Conference on
  Natural Language Processing (EMNLP-IJCNLP): System Demonstrations}, pages
  151--156, Hong Kong, China, November. Association for Computational
  Linguistics.

\bibitem[\protect\citename{Escolano \bgroup et al.\egroup }2020]{escolano:2020}
Carlos Escolano, Marta~R. Costa-juss{\`a}, Jos{\'e} A.~R. Fonollosa, and
  M\'ikel Artetxe.
\newblock 2020.
\newblock Multilingual machine translation: Closing the gap between shared and
  language-specific encoder-decoders.
\newblock {\em Corr}, abs/2004.06575.

\bibitem[\protect\citename{Firat \bgroup et al.\egroup }2016a]{firat2016multi}
Orhan Firat, Kyunghyun Cho, and Yoshua Bengio.
\newblock 2016a.
\newblock Multi-way, multilingual neural machine translation with a shared
  attention mechanism.
\newblock In {\em Proceedings of the 2016 Conference of the North {A}merican
  Chapter of the Association for Computational Linguistics: Human Language
  Technologies}, pages 866--875, San Diego, California, June. Association for
  Computational Linguistics.

\bibitem[\protect\citename{Firat \bgroup et al.\egroup }2016b]{firat2016zero}
Orhan Firat, Baskaran Sankaran, Yaser Al-Onaizan, Fatos~T. Yarman~Vural, and
  Kyunghyun Cho.
\newblock 2016b.
\newblock Zero-resource translation with multi-lingual neural machine
  translation.
\newblock In {\em Proceedings of the 2016 Conference on Empirical Methods in
  Natural Language Processing}, pages 268--277, Austin, Texas, November.
  Association for Computational Linguistics.

\bibitem[\protect\citename{Gamallo \bgroup et al.\egroup
  }2017]{gamallo2017language}
Pablo Gamallo, Jos{\'e}~Ramom Pichel, and I{\~n}aki Alegria.
\newblock 2017.
\newblock From language identification to language distance.
\newblock {\em Physica A: Statistical Mechanics and Its Applications},
  484:152--162.

\bibitem[\protect\citename{Gu \bgroup et al.\egroup
  }2019]{gu-etal-2019-improved}
Jiatao Gu, Yong Wang, Kyunghyun Cho, and Victor~O.K. Li.
\newblock 2019.
\newblock Improved zero-shot neural machine translation via ignoring spurious
  correlations.
\newblock In {\em Proceedings of the 57th Annual Meeting of the Association for
  Computational Linguistics}, pages 1258--1268, Florence, Italy, July.
  Association for Computational Linguistics.

\bibitem[\protect\citename{Johnson \bgroup et al.\egroup
  }2017]{johnson2017google}
Melvin Johnson, Mike Schuster, Quoc~V Le, Maxim Krikun, Yonghui Wu, Zhifeng
  Chen, Nikhil Thorat, Fernanda Vi{\'e}gas, Martin Wattenberg, Greg Corrado,
  et~al.
\newblock 2017.
\newblock Google’s multilingual neural machine translation system: Enabling
  zero-shot translation.
\newblock {\em Transactions of the Association for Computational Linguistics},
  5:339--351.

\bibitem[\protect\citename{Kingma and Ba}2014]{kingma2014adam}
Diederik~P Kingma and Jimmy Ba.
\newblock 2014.
\newblock Adam: A method for stochastic optimization.
\newblock {\em arXiv preprint arXiv:1412.6980}.

\bibitem[\protect\citename{Koehn \bgroup et al.\egroup }2007]{koehn2007moses}
Philipp Koehn, Hieu Hoang, Alexandra Birch, Chris Callison-Burch, Marcello
  Federico, Nicola Bertoldi, Brooke Cowan, Wade Shen, Christine Moran, Richard
  Zens, et~al.
\newblock 2007.
\newblock Moses: Open source toolkit for statistical machine translation.
\newblock In {\em Proceedings of the 45th annual meeting of the association for
  computational linguistics companion volume proceedings of the demo and poster
  sessions}, pages 177--180.

\bibitem[\protect\citename{Koehn}2005]{koehn2005europarl}
Philipp Koehn.
\newblock 2005.
\newblock Europarl: A parallel corpus for statistical machine translation.
\newblock In {\em MT summit}, volume~5, pages 79--86. Citeseer.

\bibitem[\protect\citename{Lu \bgroup et al.\egroup }2018]{lu2018neural}
Yichao Lu, Phillip Keung, Faisal Ladhak, Vikas Bhardwaj, Shaonan Zhang, and
  Jason Sun.
\newblock 2018.
\newblock A neural interlingua for multilingual machine translation.
\newblock In {\em Proceedings of the Third Conference on Machine Translation:
  Research Papers}, pages 84--92, Belgium, Brussels, October. Association for
  Computational Linguistics.

\bibitem[\protect\citename{McInnes \bgroup et al.\egroup
  }2018]{mcinnes2018umap-software}
Leland McInnes, John Healy, Nathaniel Saul, and Lukas Grossberger.
\newblock 2018.
\newblock Umap: Uniform manifold approximation and projection.
\newblock {\em The Journal of Open Source Software}, 3(29):861.

\bibitem[\protect\citename{Sennrich \bgroup et al.\egroup
  }2016]{sennrich-etal-2016-neural}
Rico Sennrich, Barry Haddow, and Alexandra Birch.
\newblock 2016.
\newblock Neural machine translation of rare words with subword units.
\newblock In {\em Proceedings of the 54th Annual Meeting of the Association for
  Computational Linguistics (Volume 1: Long Papers)}, pages 1715--1725, Berlin,
  Germany, August. Association for Computational Linguistics.

\bibitem[\protect\citename{Zoph and Knight}2016]{zoph-knight-2016-multi}
Barret Zoph and Kevin Knight.
\newblock 2016.
\newblock Multi-source neural translation.
\newblock In {\em Proceedings of the 2016 Conference of the North {A}merican
  Chapter of the Association for Computational Linguistics: Human Language
  Technologies}, pages 30--34, San Diego, California, June. Association for
  Computational Linguistics.

\end{thebibliography}
\bibliographystyle{coling}

%\section*{Appendix: Visualizations}

%\input{sections/visualization.tex}

\end{document}